\documentclass[10pt,twocolumn,letterpaper]{article}
\usepackage[norule,symbol,perpage]{footmisc}
\usepackage{icb}
\usepackage{times}
\usepackage{epsfig}
\usepackage{graphicx}
\usepackage{amsmath}
\usepackage{amssymb}
\usepackage{float}
\usepackage{xcolor}
\usepackage{amsmath,amssymb,amsfonts}

\icbfinalcopy 

\ificbfinal\fi

\makeatother

\begin{document}


\title{Gender Classification from Iris Texture Images Using a New Set of Binary Statistical Image Features.}

\author{Juan Tapia and Claudia Arellano\\
Universidad Tecnologica de Chile - INACAP \\
{\tt\small j\_tapiaf@inacap.cl}\\
\textbf{A pre-print version of the paper accepted at 12th IAPR International Conference on Biometrics. }
}

\maketitle

\begin{abstract}

Soft biometric information such as gender can contribute to many applications like as identification and security. This paper explores the use of a Binary Statistical Features (BSIF) algorithm for classifying gender from iris texture images captured with NIR sensors. It uses the same pipeline for iris recognition systems consisting of iris segmentation, normalisation and then classification. Experiments show that applying BSIF is not straightforward since it can create artificial textures causing misclassification. In order to overcome this limitation, a new set of filters was trained from eye images and different sized filters with padding bands were tested on a subject-disjoint database. A Modified-BSIF (MBSIF) method was implemented. The latter achieved better gender classification results (94.6\% and 91.33\% for the left and right eye respectively). These results are competitive with the state of the art in gender classification. In an additional contribution, a novel gender labelled database was created and it will be available upon request.
\end{abstract}

\section{Introduction}

Whenever people log onto computers, access an ATM, pass through airport security, use credit cards, or enter high-security areas, their identities need to be verified \cite{Bowyer2008281,ASH:2014:canpass}. There is tremendous interest in reliable and secure identification methods. An active research area of this involves gender classification. Algorithms for automatic gender classification have several applications. They can be used for database binning and retrieval, for intelligent user interfaces or visual surveillance. They can also be used to provide demographic information to improve social services, to facilitate payment methods and for marketing applications in general.

Gender classification based on iris images is promising despite challenging problems presented in terms of image analysis \cite{Lagree2011,Thomas2007,Tapia2013}.
The human iris is an annular part between the pupil and the white sclera. The iris has an extraordinary structure and includes many interlacing minute features such as freckles, coronas, stripes, furrows, crypts and so on. These visible features, generally called the texture of the iris, are unique to each individual \cite{Adler1965,Daugman2001,Daugman2004}. Research has also shown that the iris is essentially stable throughout a person's life. Furthermore, since the iris is externally visible, iris-based biometrics systems can be non-invasive to their users \cite{Daugman2001,Daugman2004} which is important for practical applications. All these properties (i.e., uniqueness, stability and non-invasiveness) make gender classification suitable and attractive as a complement for achieving highly reliable personal identification.

In this work a gender classification method is proposed. It uses normalised iris texture information which is codified using MBSIF. The outline of this paper is as follows: Section \ref{SOA} reviews the state of the art in gender classification methods and describes the BSIF algorithm used in this work. Section \ref{proposal} describes the pipeline of this work and the challenges faced when implementing MBSIF algorithms. Experimental set-up and the results of gender classification using several classifiers and MBSIF implementation settings are shown in Section \ref{experimentsResults}. Finally, the conclusions are presented in section \ref{conclusiones}.

\section{Related work}
\label{SOA}
\subsection{Gender Classification}

Human faces provide important visual information for gender classification \cite{ASH:2014:canpass, UIP2014}. Most work done to date on gender classification has involved the analysis of facial images and used different pattern analysis to increase the accuracy of classification \cite{He2011,Alexandre2010,Han2014,Tapia2013}.  

Previous work on gender classification from iris images has focused on handcrafted feature extraction methods using normalised NIR iris images \cite{Ojala2002,Thomas2007, Lagree2011, Bansal2012, Kannala2012,Costa-Abreu2015,Tapia2017}. Some research has utilised uniform patterns or combined uniform patterns with  non-uniform patterns to improve performance \cite{Zhou20084314,Shan2012431}. A small number of methods have used Deep Learning on Soft-biometrics such as gender with periocular NIR images \cite{KuehlkampBeckerBowyer2017, TapiaAravena2017, SinghNagpalVatsaEtAl2017}.

Tapia et al. \cite{TapiaPerezBowyer2016} classified gender directly from the same binary iris-code that is used for recognition. They found that relevant information for predicting gender is distributed across the iris, rather than localised in particular concentric bands. Therefore, selected features representing a subset of the iris region can achieve better results than when using the whole iris. They have reported 89\% correct gender prediction by fusing the best features of iris-code from left and right eyes.

Bobeldyk et al. \cite{BobeldykRoss2016} explored gender-prediction accuracy associated with four different regions from NIR iris images: the extended ocular region, the iris-excluded ocular region, the iris-only region, and the normalised iris-only region. They also used a BSIF texture operator to extract features from these four regions. The ocular region demonstrated its best performance at 85.7\%, while the normalised or unwrapped images exhibited the worst performance, with an almost 20\% decrease in performance over the ocular region. A summary of gender classification work is presented in Table \ref{my-label}.

\begin{table}[H]
\centering
\scriptsize
\caption{\label{my-label} Summary of gender classification methods using eye images. NS: Number of Subjects, I: Iris Images, P: Periocular Images, Th: Thermal, CP: Cellphone Images.}

\begin{tabular}{|c|c|c|c|c|c|}
\hline
Paper                                                                         & I/P & Source & NS & Type        & Acc. \% \\ \hline
\begin{tabular}[c]{@{}c@{}}V .Thomas et al.\cite{Thomas2007}\end{tabular}         & I   & Iris   & N/A            & NIR         & 75,00   \\ \hline
\begin{tabular}[c]{@{}c@{}}S. Lagree et al.\cite{Lagree2011} \end{tabular}         & I   & Iris   & 300            & NIR         & 62,17   \\ \hline
\begin{tabular}[c]{@{}c@{}}A. Bansal et al.\cite{Bansal2012} \end{tabular}          & I   & Iris   & 200            & NIR         & 83,60   \\ \hline
\begin{tabular}[c]{@{}c@{}}J. Tapia et al.\cite{JuanE.Tapia2014}\end{tabular}          & I   & Iris   & 1,500          & NIR         & 91.00   \\ \hline
\begin{tabular}[c]{@{}c@{}}M. Fairhurst et al.\cite{Costa-Abreu2015}\end{tabular}      & I   & Iris   & 200            & NIR         & 89,74   \\ \hline
\begin{tabular}[c]{@{}c@{}}J. Tapia et al.\cite{TapiaPerezBowyer2016}\end{tabular}          & I   & Iris   & 1,500          & NIR         & 89,00   \\ \hline
\begin{tabular}[c]{@{}c@{}}D. Bobeldyk et al.\cite{BobeldykRoss2016}\end{tabular}        & I/P & Iris   & 1,083          & NIR         & 85,70   \\ \hline
\begin{tabular}[c]{@{}c@{}}Kuehlkamp et al.\cite{KuehlkampBeckerBowyer2017}\end{tabular}        & I/P & Iris   & 1,500          & NIR         & 80.00   \\ \hline
\begin{tabular}[c]{@{}c@{}}J. Tapia.\cite{Tapia2017}\end{tabular}        & I/P & Iris   & 1,500          & NIR         & 79.33  \\ \hline
\begin{tabular}[c]{@{}c@{}}J. Tapia et al.\cite{TapiaAravena2017}\end{tabular}        & I & Iris   & 1,500          & NIR         & 83.00  \\ \hline
\begin{tabular}[c]{@{}c@{}}J. Merkow et al.\cite{Merkow2010}\end{tabular}         & P   & Faces  & 936            & VIS         & 80,00   \\ \hline
\begin{tabular}[c]{@{}c@{}}C. Chen et al.\cite{ChenRoss2011}	\end{tabular}           & P   & Faces  & 1,003          & NIR/Th & 93,59   \\ \hline
\begin{tabular}[c]{@{}c@{}}Castrillon-Santana et al. \cite{Castrillon-SantanaLorenzo-NavarroRamon-Balmaseda2016}\end{tabular} & P   & Faces  & 1,500  & VIS & 92,46 \\ 
\hline
\begin{tabular}[c]{@{}c@{}}Rattani et al.\cite{RattaniReddyDerakhshani2017}\end{tabular} & P   & Iris  & 550  & VIS/ CP & 91,60 \\ 
\hline
\begin{tabular}[c]{@{}c@{}} J. Tapia et al.\cite{TapiaViedma2017} \end{tabular} & P   & Iris  & 120/120  & NIR/VIS & 90,00 \\ 
\hline
\end{tabular}
\end{table}

\subsection{Binary Statistical Image Feature (BSIF) }
\begin{figure*}[h]
\begin{centering}
\begin{tabular}{cc}
\includegraphics[scale=0.25]{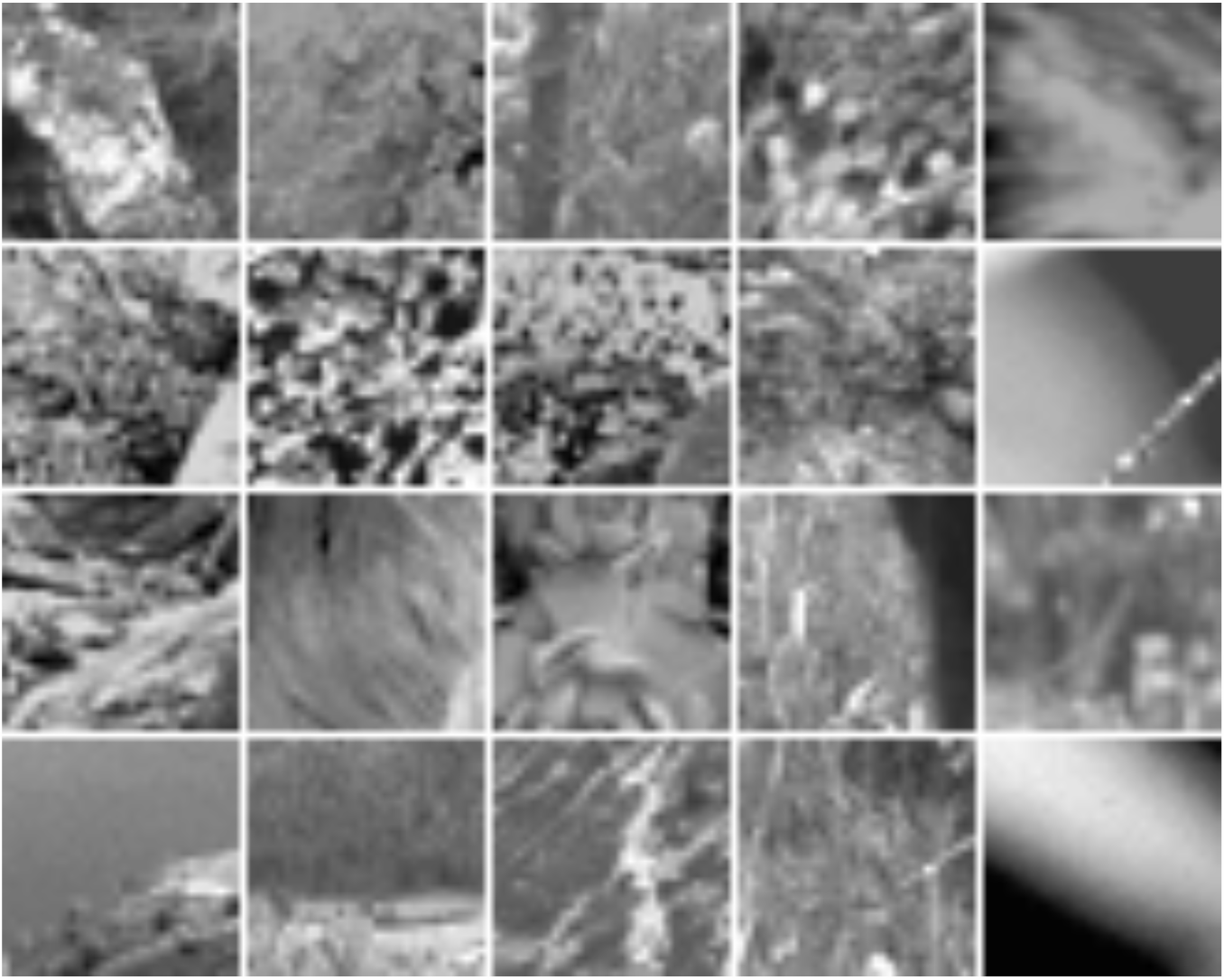} & \includegraphics[scale=0.25]{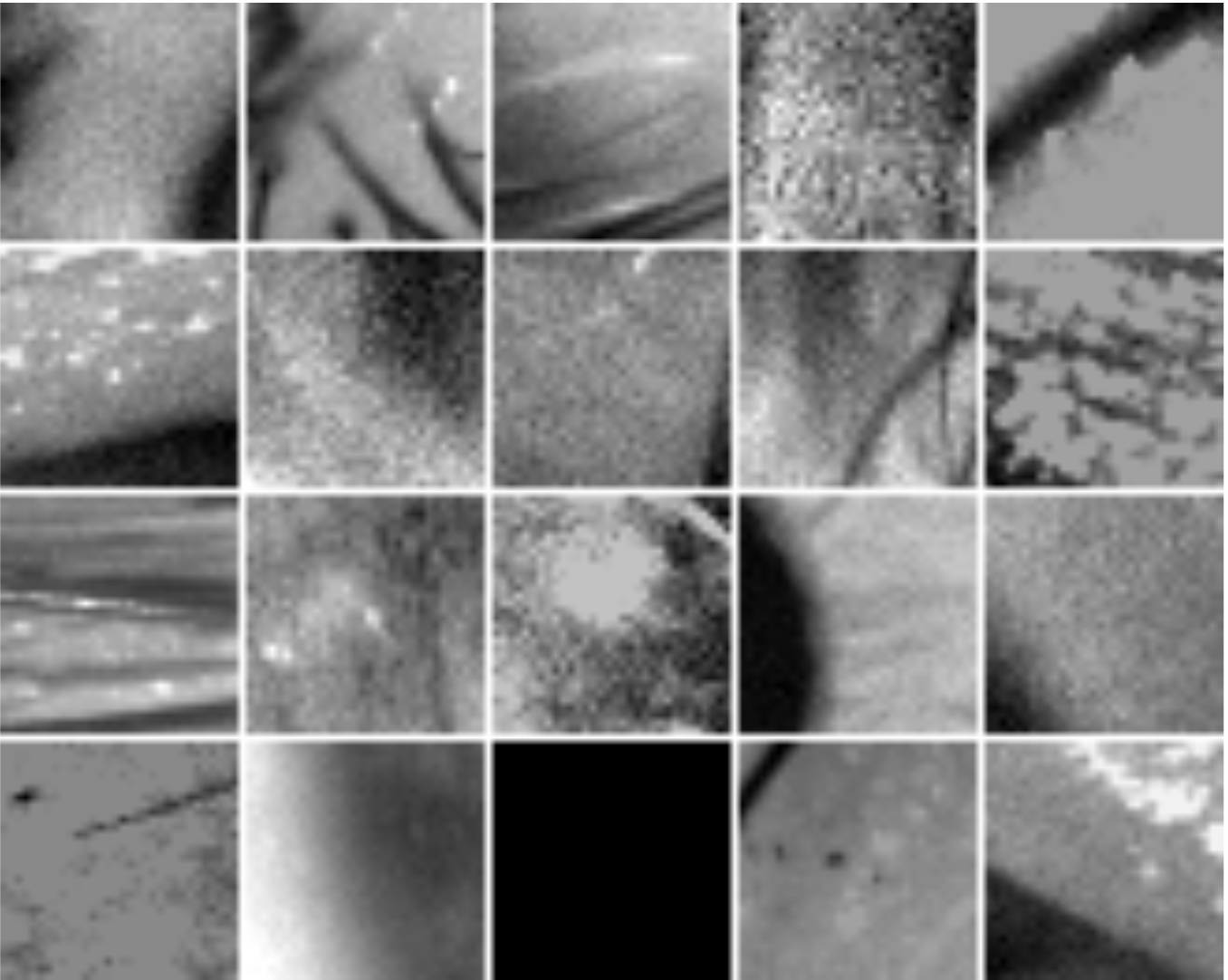}\tabularnewline
\end{tabular}
\par\end{centering}

\caption{ \label{filter} Left, Example of patches extracted from natural images for traditional BSIF. Right, Example of patches extracted from Eye images for Modified BSIF.}
\end{figure*}

BSIF \cite{Kannala2012} is a local descriptor constructed by binarising the responses to linear filters. In contrast to previous binary descriptors, \textbf{\emph{the filters learn from thirteen natural images}} using independent component analysis (ICA). The code value of pixels is considered as a local descriptor of the image intensity pattern in the pixels' surroundings. The value of each element (i.e bit) in the binary code string is computed by binarising the response of a linear filter with a zero threshold. Each bit is associated with a different filter, and the length of the bit string determines the number of filters used. The set of filters is learned from a training set of natural image patches by maximising the statistical independence of the filter responses \cite{Hyvrinen}(See Figure \ref{filter}). The details of the parameters learned by the linear filters are described below:
Given an image patch $X$ of size $l\times l$ pixels and a linear filter $W_{i}$ of the same size, the filter responses $s_{i}$ are obtained by:
\begin{equation}
s_{i}=\sum_{u,v}W_{i}(u,v)X(u,v)=w_{i}^{T}x,
\end{equation}

Where, vector notation is introduced in the latter stage, for instance the vector $w$ and $x$ contain the pixels of $W_{i}$ and $X.$ The binarised feature $b_{i}$ is obtained by setting $b_{i}=1$ if
$s_{i}>0$ and $b_{i}=0$ otherwise. Given $n$ linear filters $W_{i}$, we may stack them to a matrix $W$ of size $n \times l^{2}$ and compute all responses at once, i.e. $s=Wx$.  We obtain the bit string $b$ by binarising each element $s_{i}$ of $s$ as above. Thus, given the linear feature detectors $W_{i}$, computation of the bit string $b$ is straightforward. Also, it is clear that the bit strings for all image patches of size $l\times l$, surrounding each pixel of an image can be computed conveniently by $n$ convolutions. 

The final image is obtained by:

\begin{equation}
CodeIM=CodeIm+(Cr>0)*(2^{nbits})
\end{equation}

Where, $CodeIm$ is an accumulative image, $Cr$ is the convolution between the filter and the image that is later binarised and multiplied by the number of bits. For instance, if we use 9 bits then we compute
$CodeIm$ for $2^{1}$ later for $2^{2}$ up to $2^{9}$. The final image will be the sum of the 9 images for each $CodeIm$. 

BSIF have been used for several applications including biometrics from iris images \cite{KomulainenHadidPietikaeinen2014,DoyleBowyer2015,RathgebStruckBusch2016}. In this work, a gender classification algorithm using normalised NIR iris images is proposed. It uses a similar pipeline than iris recognition systems. The iris is segmented and occlusions are masked.
BSIF can be sensitive to image boundaries and the occlusion mask creating artificial texture which may mislead gender classification results. 

This paper explores a new set of filters (See Figure \ref{filter}) trained from thirteen eye images instead of natural images as used in traditional approach. The influence of the filter size, the padding (boundaries) and the number of bits used when implementing MBSIF algorithm are also explored.

\section{Gender classification using BSIF}
\label{proposal}
This paper proposes the use of the same pipeline that is used for iris recognition systems. The input image is segmented in a pre-process step. The iris region is then transformed to a polar space and codified using MBSIF. Finally, gender classification is performed using a new database and several classifiers (Section \ref{classifier}).

\subsection{Iris Segmentation and Normalisation}

\begin{figure}[h]
\begin{centering}
\begin{tabular}{|c|c|c|}
\hline 
 & Right Eye& Left Eye\tabularnewline
\hline 
a) & \includegraphics[scale=0.12]{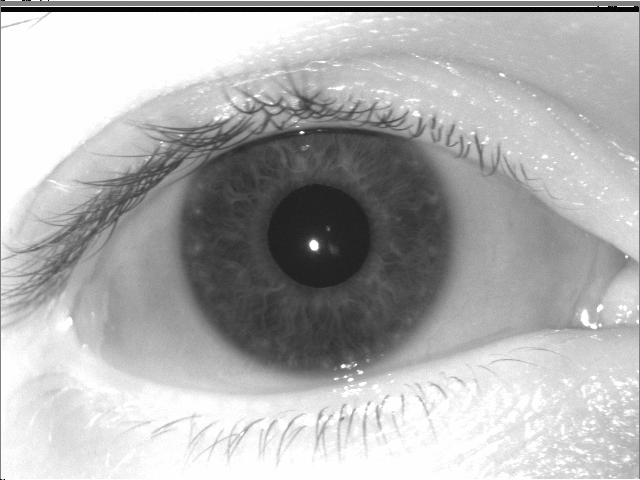} & \includegraphics[scale=0.12]{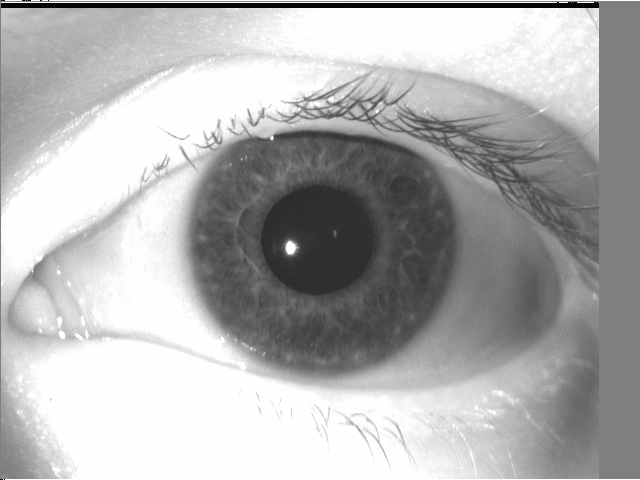}\tabularnewline
\hline 
b) & \includegraphics[scale=0.12]{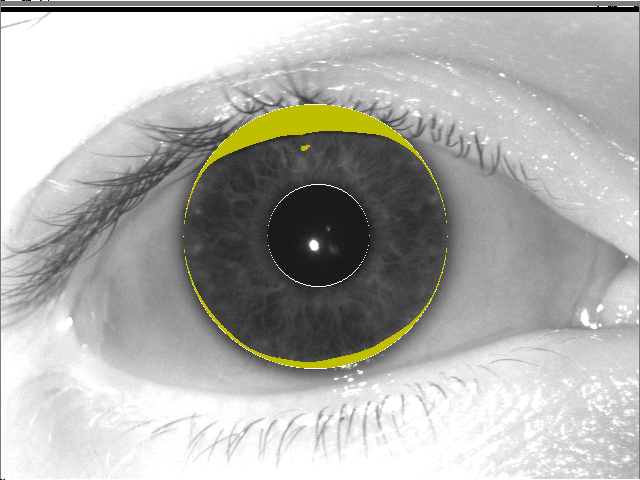} & \includegraphics[scale=0.12]{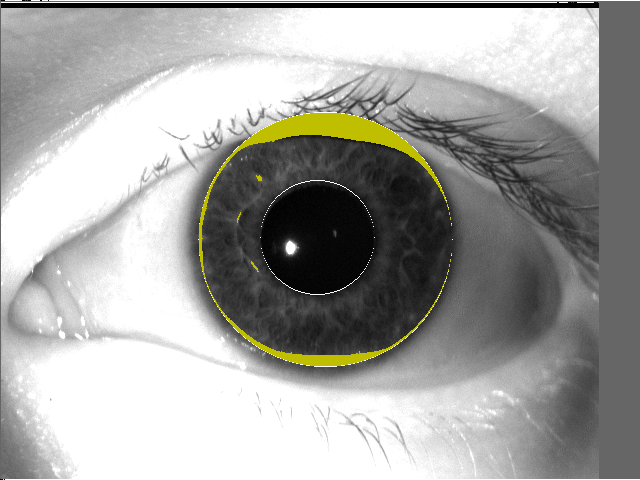}\tabularnewline
\hline 
c) & \multicolumn{2}{c|}{\includegraphics[scale=0.8]{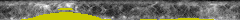}}\tabularnewline
\hline 
 d) & \multicolumn{2}{c|}{\includegraphics[scale=0.8]{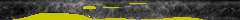}}\tabularnewline
\hline 
\end{tabular}
\par\end{centering}
\caption{\label{pipeline1} Two original images from right and left eye (a). Segmented and masked images with eyelid and eyelash detection using Osiris (b). Images (c) and (d) are normalised images from the right and left eye both with the mask in yellow.}
\end{figure}

The iris is detected from the input image using commercial software Osiris \cite{osiris}. A segmentation mask occludes the eyelids, eyelashes and specular reflection  portions of the iris image which are not useful for gender classification.  It is important to note that iris images of different persons, or even the left and right iris images for a given person, may not present exactly the same mask and imaging conditions (see Figure \ref{pipeline1}). Illumination by LEDs during capture may come from either side of the sensor, specular highlights may be present in different places in the image. Eyelid and head position may also affect segmentation.

The segmented iris is normalised or unwrapped with radial $(r)$ and angular $(\theta)$ resolutions which determine the size of the rectangular iris image.  The size of the normalised iris can significantly influence the iris recognition rate. In this work, a rectangular image of $20(r)$x $240(\theta)$ created using Osiris software \cite{osiris} with automatic segmentation is used for all experiments.

\subsection{BSIF filters application}

BSIF filters compute the convolution with each normalised masked image. 
Each filter represents a different pattern. The final image is the results of all previous images binarised by $2^n$ bits. The best filter size is one that represents the correct size of the mask with the lowest number of bits. If the filter is smaller than the mask, then artificial texture information will be created and the resulting image will not well represent its original information. On the other hand, if the mask of the iris is larger than the filter, a flat area will be obtained and the filter will need to be adjusted by reducing its size.
Since the size of the normalised iris image is 20 $\times$ 240, special care needs to be taken in order to minimise the effects of boundary and its influence on filter size.
A common approach to dealing with border effects is to pad the original image with extra rows and columns based on your filter size.

\begin{figure*}[h]

\begin{centering}
\begin{tabular}{ccc}
\\

(a) & \includegraphics[scale=0.3]{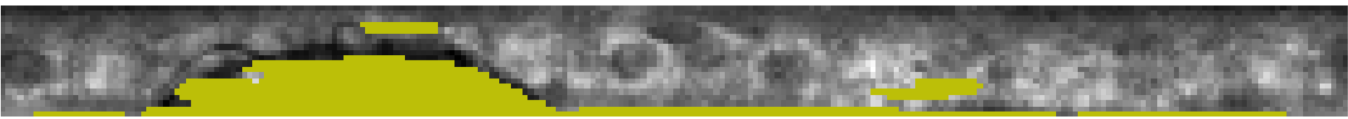}& \includegraphics[scale=0.3]{Images/751_eye_left_raw}
\tabularnewline
\\
(b) & \includegraphics[scale=0.31]{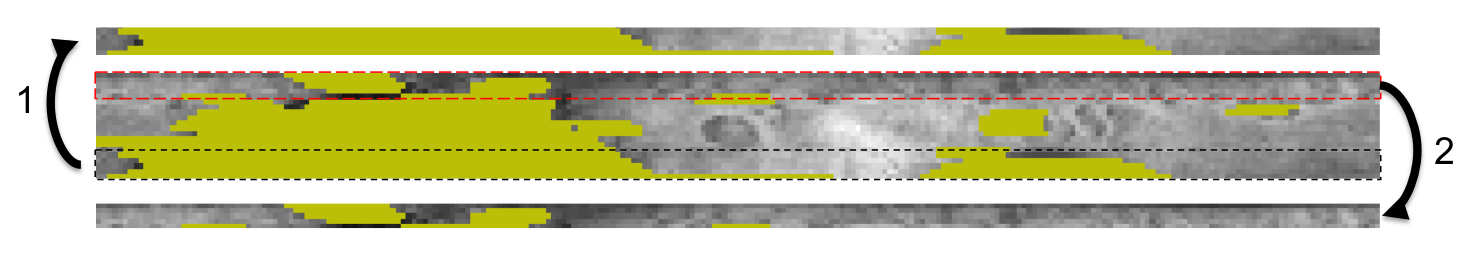}& \includegraphics[scale=0.31]{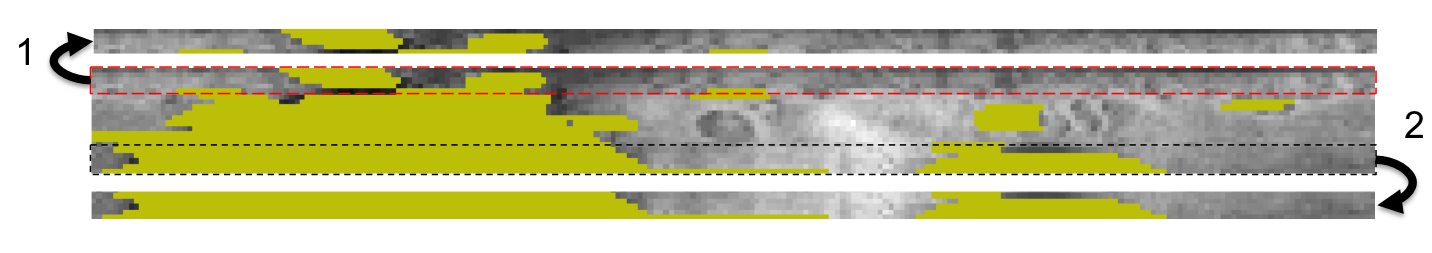}
\tabularnewline
\\
(c) & \includegraphics[scale=0.31]{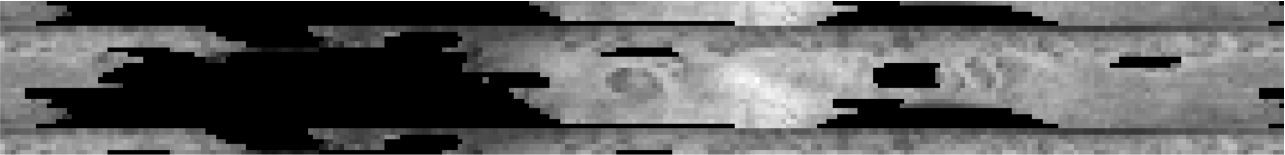}& \includegraphics[scale=0.32]{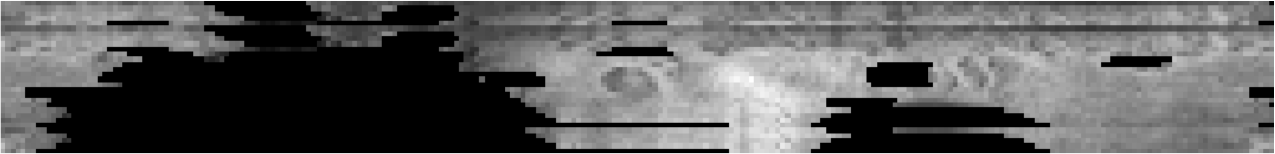}
\tabularnewline
\\
(d) & \includegraphics[scale=0.31]{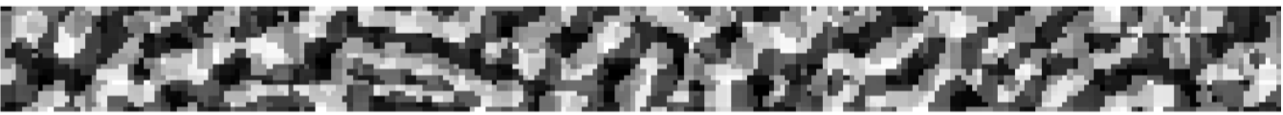}& \includegraphics[scale=0.32]{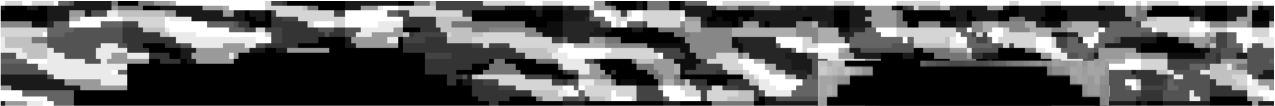}
\tabularnewline
\\
&(A)&(B)\\
&Traditional BSIF Implementation & Proposed MBSIF Implementation \\
\end{tabular}
\par\end{centering}
\caption{\label{figure_BSIF} Column (A) shows an example of traditional BSIF implementation. (a) corresponds to the input normalised iris image with the mask information in yellow. (b) illustrates padding implemented were bands 1 and 2 are wrapped on the image and (c) the resulting image after applying BSIF filters (11x11 pixels and 9 bits). A similar example is shown in column (B). In this case two bands of pixels (1 and 2) are replicated at the top and bottom of the image. The resulting image in (d) replicates the mask of the input image without adding extra artificial texture}
\end{figure*}

Traditional implementation of BSIF increases the size of the image and wraps the filter around it.
Unfortunately, this implementation directly affects the results of the binarised iris image. Figure \ref{figure_BSIF} (A) shows an example where this implementation is used. The first row (a), shows the normalised iris image obtained directly from Osiris software \cite{osiris}. The second row (b) shows the extra rows added through the wrapping process. A $5\times240$ pixel band is added to the top and bottom of the original image. Additional bands of $5\times20$ pixels are added to the vertical sides of the image (left and right). Note that the horizontal band added to the top of the image represents the bottom of the original image (mask area) and, the horizontal band added to the bottom of the image represents the top of the original image (Figure \ref{figure_BSIF}, column (A), row (b)). This implementation directly affects the resulting binarised image since the boundary added creates artificial texture as can be seen in the resulting images in Figure \ref{figure_BSIF}, column (A), row (d). 

A alternative way to deal with border effects is to pad the original image with zeros (Or a constant value), reflecting the image at the borders or replicating the first and last row/column as many times as needed

In order to overcome the boundary effect of traditional BSIF implementation a portion of the image is replicated in both directions (top and bottom). Figure \ref{figure_BSIF}, column (B), shows the details of this implementation and their effect on binarised images. As can be seen in column (B) row (d), the resulting binarised image follows the pattern of the input masked image. Therefore, there is no extra information artificially added to the iris image. This approach should better represent the information contained in the input images. A new set of filters were obtained by using a novel set of eyes images (instead of natural ones). These images were used to extract patches and to train our modified version of the algorithm (MBSIF). In the experimental section several filters size are tested and compared. Two approaches are implemented, MBSIF and MBSIF histogram.

\subsection{Gender classification}
\label{classifier}
Several classification algorithms are used to test gender information from iris texture images. Those algorithms are: Adaboost M1, LogitBoost, GentleBoost, RobustBoost, LPBoost, TotalBoost and RusBoost.
Additionally, a Random Forest classifier with 500 trees, a Gini Index, and a LIBSVM classifier with Gaussian Kernel (RBF) were also used. A comparison of the results obtained with these classifiers is shown in section \ref{experimentsResults}. 
\subsection{Databases}
\textbf{GFI-UND:} The GFI-UND database used in this paper contains images taken with an LG 4000 sensor. This dataset is the same used in \cite{TapiaPerezBowyer2016}. The LG 4000 uses near-infrared illumination and acquires 480x640, 8-bit/pixel images. Examples of iris images are shown in Figure \ref{pipeline1}. The GFI-UND iris database was used to train and test a gender classifier.

For each subject (750 males and 750 females, for a total of 3,000 images), one left eye image was selected at random from the set of left eye images, and one right eye image was selected at random from the right eye images. A training portion of the dataset was created by randomly selecting 80\% of the males and 80\% of the females. A 5-fold cross-validation on this training set is used to select parameters for each classifier. Once the parameter selection was finalised, a classifier was trained on the full 80\% of the training data, and a single evaluation was made on 20\% of the test data.  Experiments are conducted separately for the left iris and the right iris. The masks were set to zero in all images.
To the authors' understanding, the GFI-UND database  \cite{TapiaPerezBowyer2016} is the only dataset created exclusively for gender classification from iris images. It it a person-disjoint set with 1,500 different subjects. 

\textbf{UNAB-Val:} As an additional contribution, a new gender-labelled database was created. This is a person-disjoint dataset that was captured using an iCAM TD-100 NIR sensor. The iCAM TD-100 uses near-infrared illumination and acquires 480x640 8-bit pixels per image. This set of iris images were obtained over 5 sessions with 66 female and 70 male subjects. Each subject has 5 images per eye. In total 660 female images and 700 male images were captured. This database is to be increased continuously since the capturing process is active as of writing. This database will be available upon request. Additional datasets were requested but unfortunately were not available \cite{KumarPassi2010,SunTan2009,RattaniDerakhshaniSaripalleEtAl2016}.
\section{Experiments and results}
\label{experimentsResults}
\begin{figure}[ht]
\centering
\includegraphics[scale=0.26]{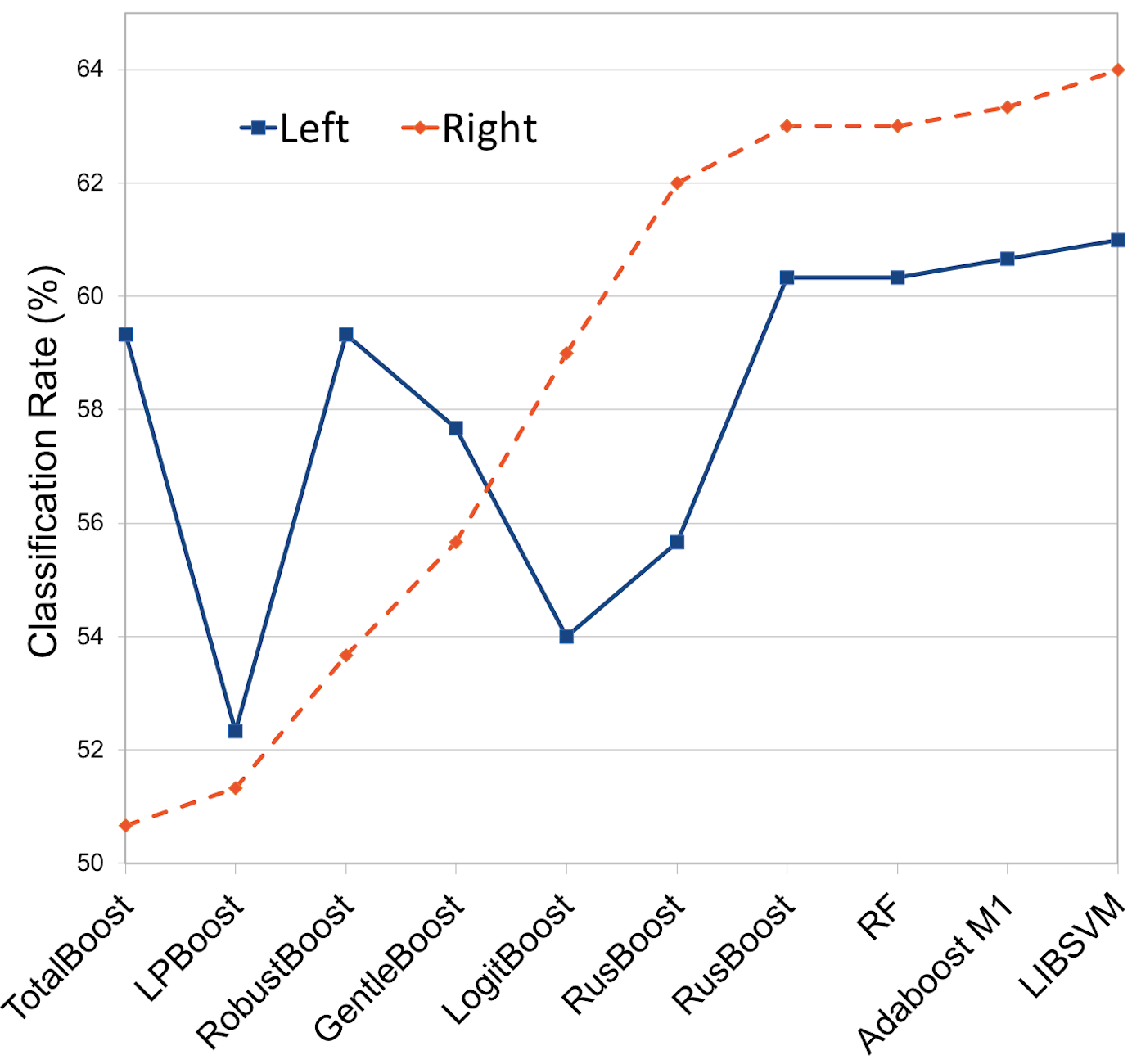}\\
\caption{\label{Classifiers}Classification rates for the left and right eye when using several classifiers and standard BSIF implementation.}
\end{figure}

Several experiments were performed in order to test the use of MBSIF for gender classification. Figure \ref{Classifiers} shows gender classification results when using the left and right eye data set (from GFI-UND database) and 10 different classifiers. In this experiment, the BSIF algorithm was implemented using the standard padding as shown in Figure \ref{figure_BSIF} (A). The best classifiers for both eyes are Adaboost and SVM. Several filter sizes (from 5 $\times$5 up to 13$\times$13) and  number of bits from 5 to 12 were used. The best results are shown in Figure \ref{Classifiers}. They were achieved when a filter size of 13 $\times$ 13 and 8 bits was used for the left eye images and a filter of 13 $\times$ 13 and 7 bits for the right eye. The maximum classification rate obtained with this implementation (BSIF) was $65\%$ and $67\%$ for the left and right eye respectively. 

\begin{figure*}[htb]
\begin{centering}
\begin{tabular}{cc}
Left Eye Images & Right Eye Images \\

\includegraphics[scale=0.32]{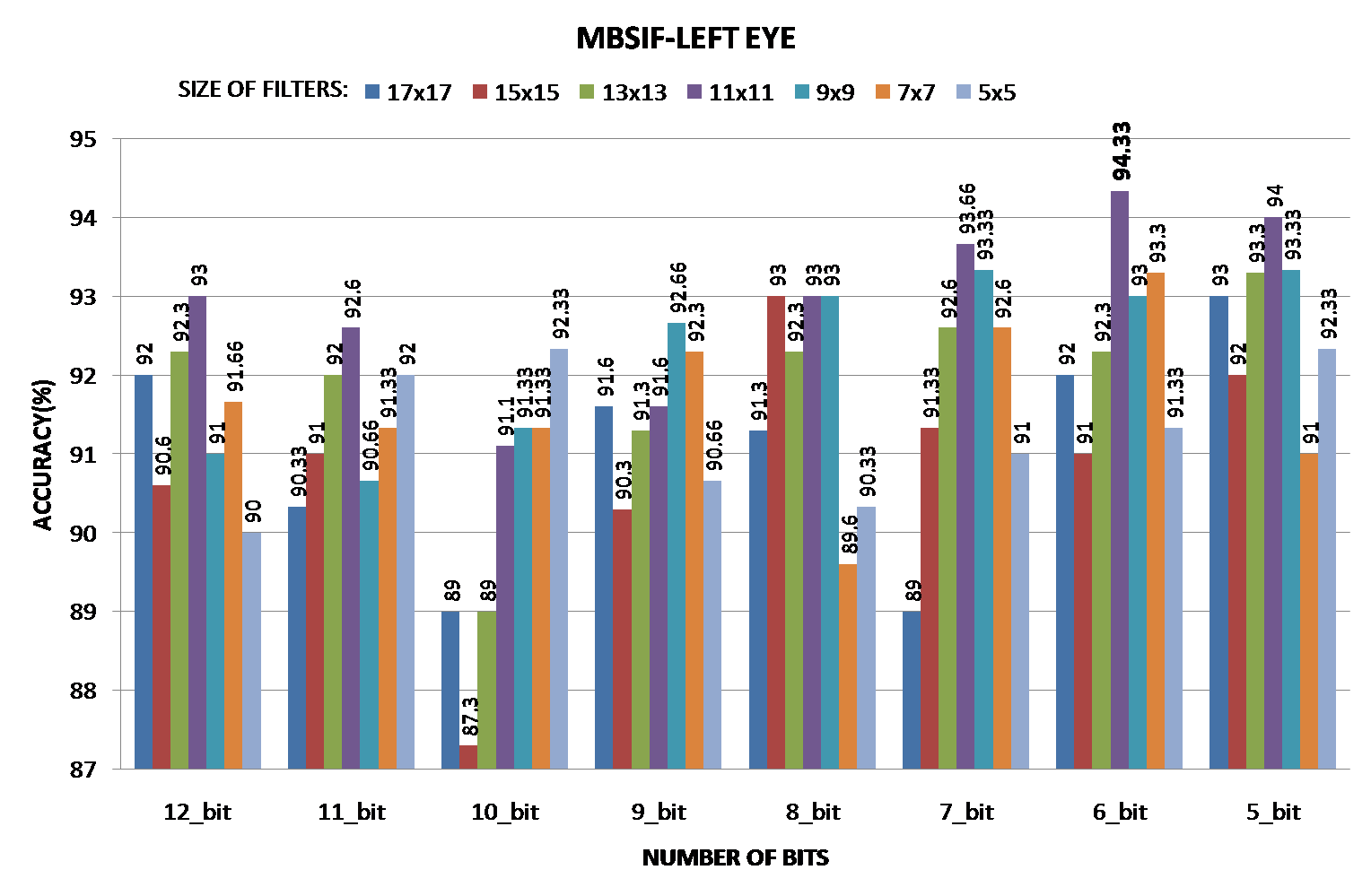}&\includegraphics[scale=0.32]{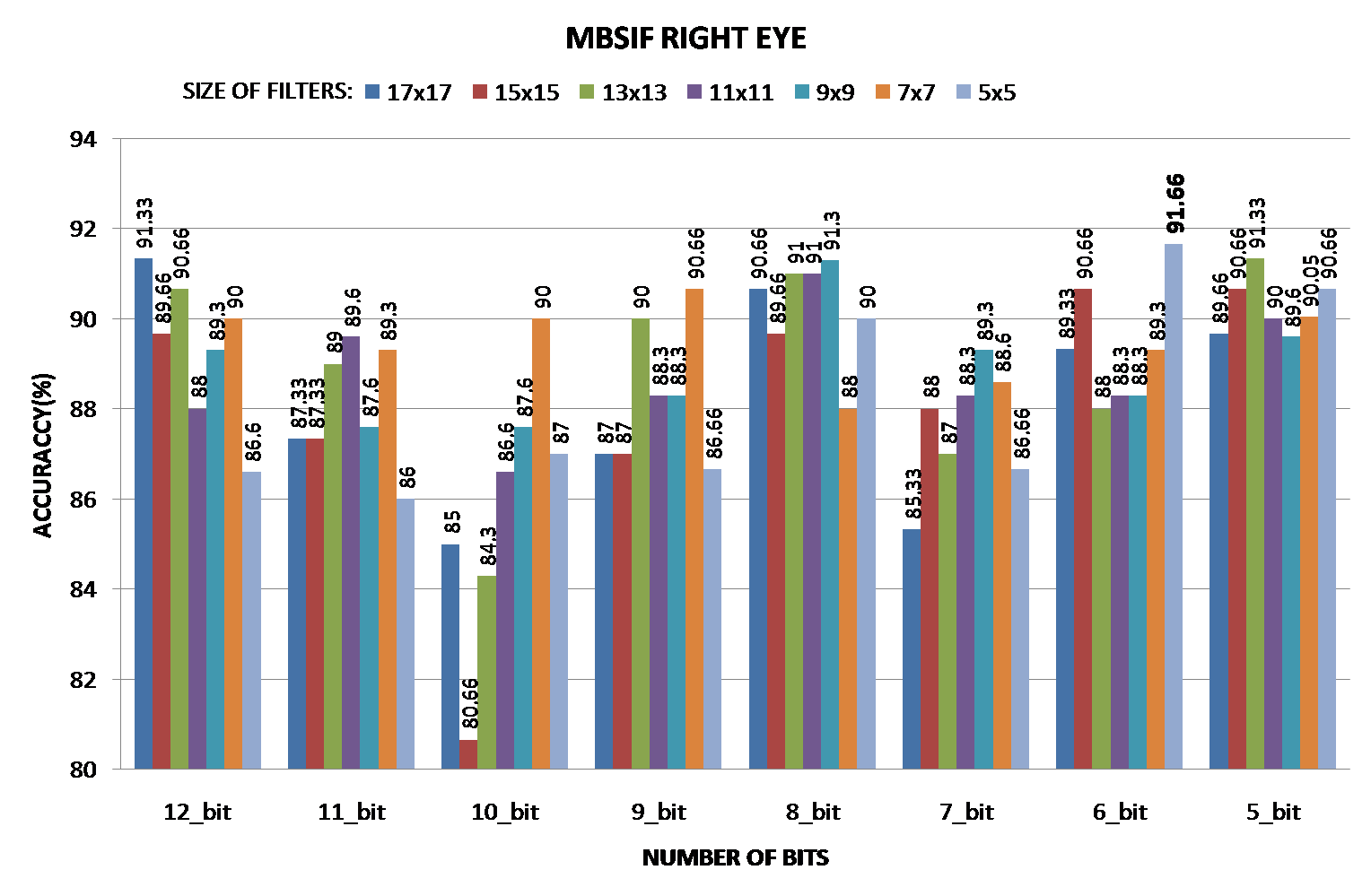}\\
     (a) & (b)\\
   \includegraphics[scale=0.33]{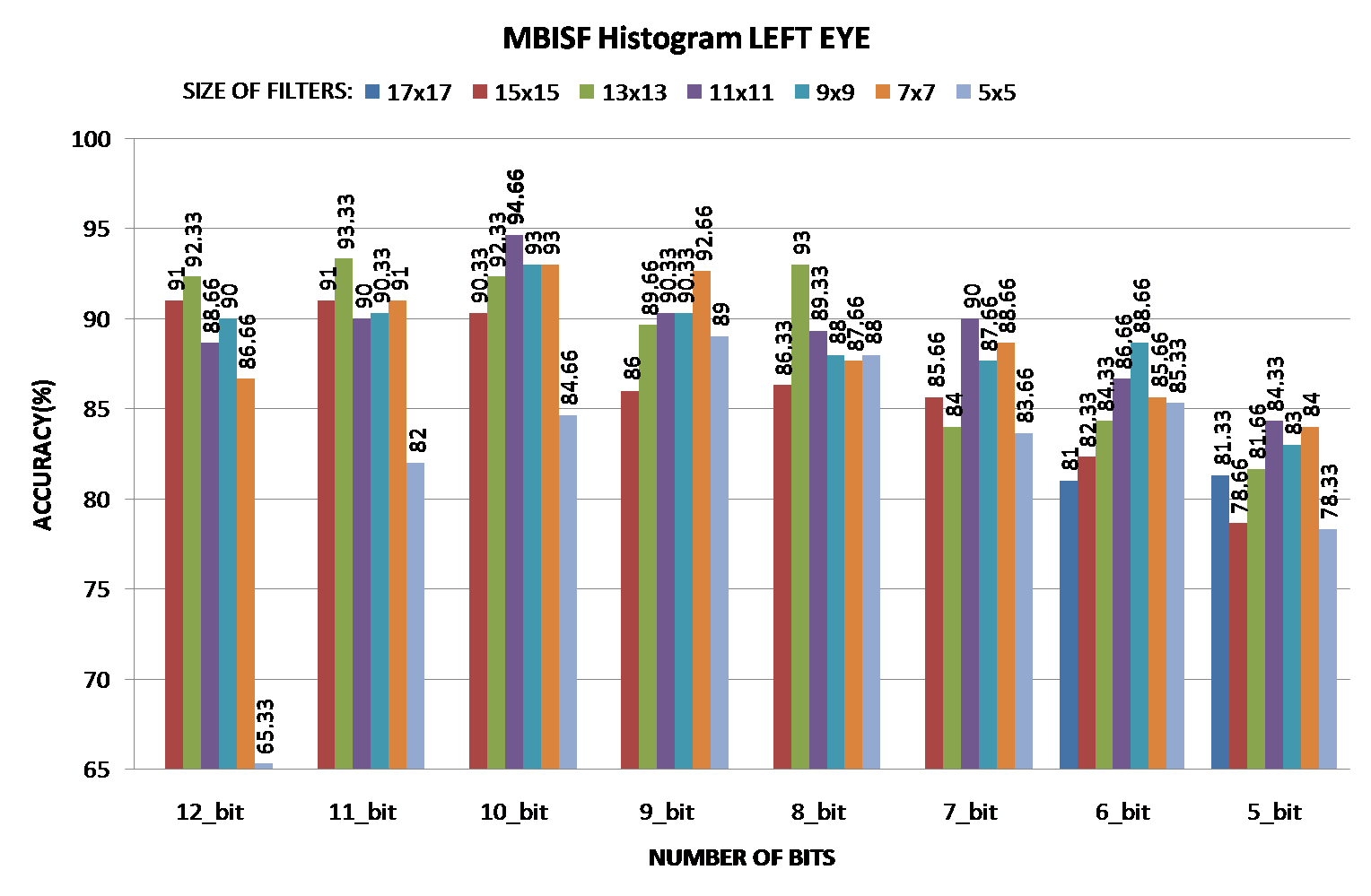}  & \includegraphics[scale=0.33]{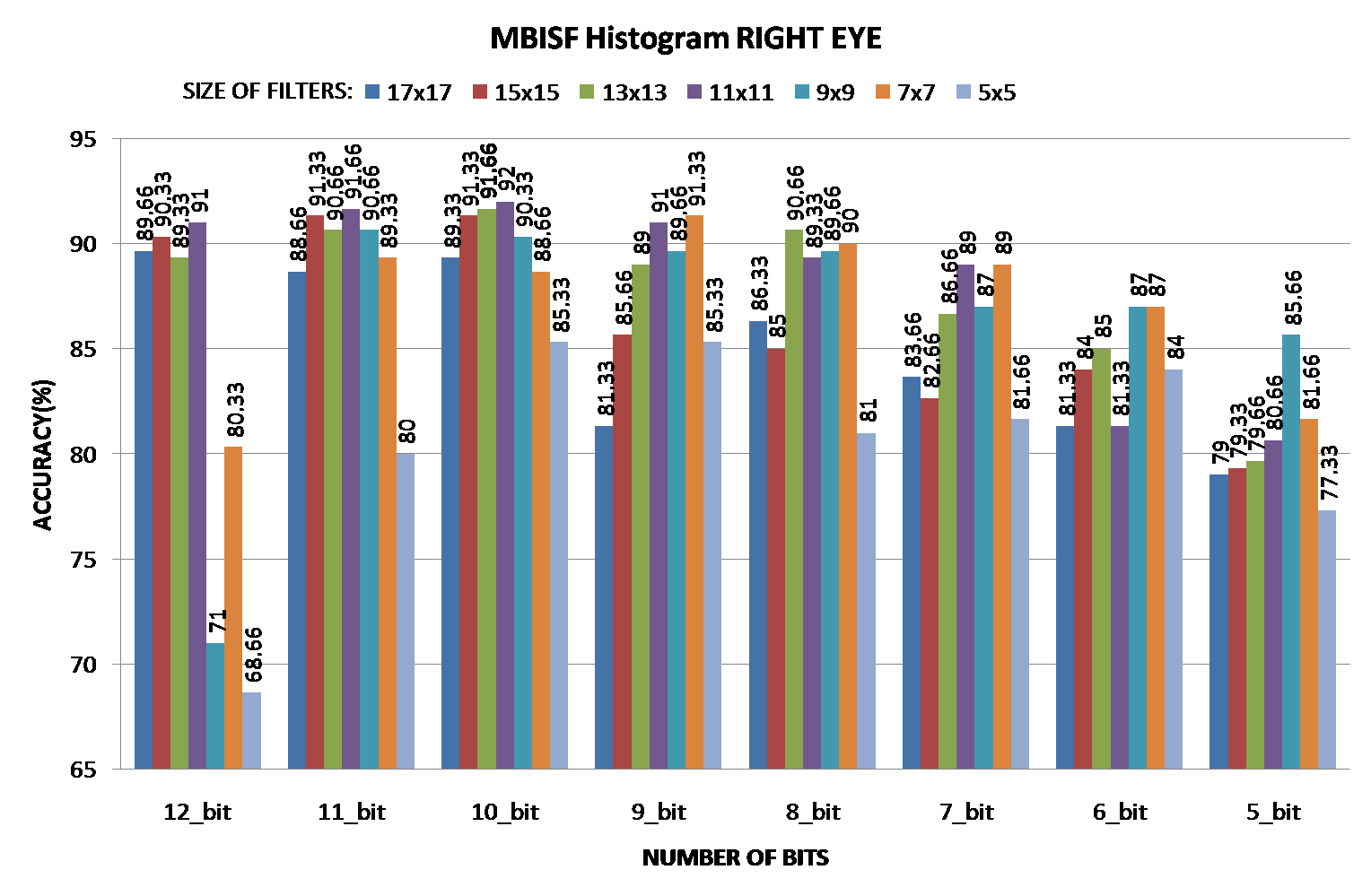} \\
     & \\
     (c) & (d)\\
   
\end{tabular}

\par\end{centering}

\caption{\label{graficos} Gender classification results using MBSIF and BSIF histogram for left and right eye images when using different filters size and number of bits from 5 up to 12.}
\end{figure*}

In order to find the best classification rate with our proposed MBSIF algorithm, several filter sizes (5x5, 7x7, 9x9, 11x11, 13x13, 15x15 and 17x17) with a different number of bits (from 5 bits up to 12 bits) were tested. The number of bits represent the number of filters used in the convolution. Experiments using the entire image (all the filter sizes and from 5-12 bit) and using the normalised histogram of images were performed (See Figure \ref{graficos}).
One of the advantages of using the normalised histogram is that the vector size of each image is smaller. It only depends on the number of bits. For instance, when using 5 bits, the resulting vector has 32 bins, whereas when using 6 bits, the resulting vector has 64 bins and so on.
Figure \ref{graficos} shows results for the left and right eye images using our proposed implementation of BSIF for both cases: when using the entire image and when using the histogram. In the case of left eye images the best result (94.33\%) was obtained with the filter 11x11 and 6 bits. This represents 144  correct identifications out of 150 male images and 140 correct identifications out of 150 female images. A slightly improved result was achieved when using the MBSIF histogram (94.66\%). In this case, the best result was obtained with a 11x11 filter and 10 bits (1024 bins). For right eye images the best results when using the proposed MBSIF implementation was $91.66\%$ and it was obtained with a 11x11 filter and 10 bits (2,048 bins). Gender classification results were slightly better when using the MBSIF histogram ( 92.00\%). In this case, results represent 140 out of 150 for male and 136 out of 150 for female images.

A summary of the best results obtained from the experiments is shown in Table \ref{tablafinal}. The best gender classification rates were achieved when boundaries of the normalised iris texture were replicated (Figure \ref{figure_BSIF}(B)) instead of wrapped around (Figure \ref{figure_BSIF}(A)). The algorithm was trained using the GFI-UND database and tested using the GFI-UND-Val and UNAB-Val datasets. The difference was only 4\% on average with both datasets.
\begin{table}[H]
\centering
\scriptsize
\caption{\label{tablafinal}Summary of gender classification rates using BSIF, MBSIF and MBSIF histogram. FS: Filter Size, NB: Number of bits.}
\begin{tabular}{|c|c|c|c|c|c|}
\hline 
Method   &FS& NB &Database&Left-Eye &Right-Eye  \\ 
&&&&(\%)&(\%)\\ \hline \hline
BSIF (A) & 11$\times$11 &12&GFI-UND&$61.67$&$67.00$ \\ \hline
MBSIF (B)& 11$\times$11 &6&GFI-UND&$94.33$&$91.66$ \\ \hline
MBSIF-H (B)& \textbf{11$\times$11} &\textbf{10}&GFI-UND&$\textbf{94.66}$&$\textbf{92.00}$ \\ \hline
\end{tabular}
\end{table}

\vspace{-0.5cm}
\section{Conclusions}
\label{conclusiones}

BSIF filters can extract and encode general patterns present in traditional images such as faces or periocular images but when applied to normalised iris images with masks, artificial textures are produced.  These artificial textures can affect gender classification rates.  Through this paper experiments have shown that special care needs to be taken on boundaries when dealing with BSIF filters. 

The patterns detected by traditional BSIF method do not represent the texture of the iris well. Traditional BSIF use thirteen natural images to create the filter patches. The filter created from Eye-Images was more suitable to capture the texture inside the iris. This also allows the gender classification rate to be improved. 

Traditional setting of BSIF increases the image size by wrapping the image values. This implementation has an impact on gender classification rates when using masked normalised iris images. Under this setting, gender classification rates of only 61\% and 67\% were achieved for right and left eye images respectively. In order to overcome the boundary effect of traditional BSIF implementation a portion of the image is replicated in both directions (top and bottom). This implementation improved the gender classification result considerably up to 94\% and 92\% for left and right eye images respectively. The best results were achieved when the MBSIF histogram was used.
There are clear computational advantages to predicting gender from the normalised image rather than computing another different texture representation. This method can be easily included in the same pipeline of recognition systems. The use of the normalised iris can reduced computational cost thanks to the small size of the image. This is particularly important when large amounts of data needs to be processed such as gender classification in highly populated countries (i.e India, china).
Experiments were validated using two databases and several classifiers. Gender classification results obtained were competitive with the state of the art. As an additional contribution,  a new gender-labelled database was created and will be available to other researchers upon request.  

\vspace{-0.3cm}
\section{Acknowledgments}
\vspace{-0.05cm}
This research was partially funded by FONDECYT INICIACION 11170189   and Universidad Andres Bello, DCI.

{\small
\bibliographystyle{ieee}
\bibliography{References_OC}
}

\end{document}